\documentclass[wcp]{jmlr}

\usepackage{longtable}

\usepackage{booktabs}

\usepackage{amsmath,amssymb,amsfonts}
\usepackage{graphicx}
\usepackage{xcolor}
\usepackage{color}
\usepackage[utf8]{inputenc}
\usepackage[english]{babel}
\usepackage{caption,subcaption}
\usepackage{algorithmic}
\usepackage{algorithm}
\usepackage{multirow}

\newcommand{\fw}{0.45}

\newcommand{\fcontroller}{0.45}
\renewcommand{\columnwidth}{\textwidth}
\title[ReSet: Recurrent Dynamic Routing in ResNets]{ReSet: Learning Recurrent Dynamic Routing in ResNet-like Neural Networks}

\author{\Name{Iurii Kemaev} \Email{y.kemaev@gmail.com}\\
  \addr Skolkovo Institute of Science and Technology, Moscow, Russia\\
  \addr National Research University Higher School of Economics, Moscow, Russia\\
 \AND
\Name{Daniil Polykovskiy} \Email{daniil.polykovskiy@gmail.com}\\
  \addr Moscow State University, Moscow, Russia\\
  \addr National Research University Higher School of Economics, Moscow, Russia
\AND
\Name{Dmitry Vetrov} \Email{vetrovd@yandex.ru}\\
\addr National Research University Higher School of Economics, Moscow, Russia
}

\begin{document}

\maketitle

\begin{abstract}
Neural Network is a powerful Machine Learning tool that shows outstanding performance in Computer Vision, Natural Language Processing, and Artificial Intelligence. In particular, recently proposed ResNet architecture and its modifications produce state-of-the-art results in image classification problems. ResNet and most of the previously proposed architectures have a fixed structure and apply the same transformation to all input images. 

In this work, we develop a ResNet-based model that dynamically selects Computational Units (CU) for each input object from a learned set of transformations. Dynamic selection allows the network to learn a sequence of useful transformations and apply only required units to predict the image label. We compare our model to ResNet-38 architecture and achieve better results than the original ResNet on CIFAR-10.1 test set. While examining the produced paths, we discovered that the network learned different routes for images from different classes and similar routes for similar images. 
\end{abstract}

\section{Introduction}
Recent advances in Neural Networks made them prevalent in many problems of computer vision like image denoising, segmentation and captioning. Recently, they achieved and even surpassed a human-level performance in a wide range of tasks, including image classification and object detection~\cite{he2016deep, russakovsky2015imagenet}. Residual learning~\cite{he2016deep} is a building block for most of state of the art models. Deep Residual Networks showed outstanding results on the ILSVRC2015 challenge and quickly became one of the most promising architectures for many efficient models in computer vision.

Residual learning is based on the idea of shortcut connections between layers. If a standard layer learns some parametric mapping $y = f(x)$ of an input $x$, residual learning suggests to learn the mapping in a form of $y = x + f(x)$. The function $f$ usually takes a form of a few convolutional layers with nonlinearities in between. The form of mapping  $y = x + f(x)$ is motivated by the fact that very deep neural networks without shortcut connections often exhibit worse performance than their shallower counterparts. Counterintuitively, as shown in~\cite{he2016deep}, this result is due to underfitting, not overfitting. However, intuitively, deeper networks should not generalize worse, since last layers of the network can learn an identity mapping $y=x$, effectively learning a shallower model.  Since a network cannot learn an identity mapping $f(x)=x$, adding a shortcut connection allows the latter layers to learn an identity mapping $y=x$ by setting $f(x) = 0$, which can usually be achieved by zeroing out all the weights. The idea of residual networks paved the way for deep models consisting of hundreds~\cite{huang2017densely} and even thousands~\cite{he2016identity} of layers.

One of the most notable properties of ResNets is their robustness to deletion and reordering~\cite{veit2016residual, srivastava2015highway} of layers. Randomly deleting or reordering of multiple layers of the trained ResNet does not significantly decrease the performance of the model. However,  quality of non-residual models, like the VGG family~\cite{simonyan2014very}, drops to the random prediction after these operations. This leads to the hypothesis that models with ResNet-based architectures exhibit ensemble-like behavior in the sense that their performance smoothly correlates with the number of layers.~\cite{veit2016residual} considered this effect as the result of an exponentially large number of paths of different lengths formed by skip connections. Although these paths are trained simultaneously, they are weakly dependent, and deleting some layers of the network affects only a small number of paths. From this point of view, authors proposed to treat ResNets as an exponential ensemble of shallow networks.

While studying this notable property, we assume that for each input object, only a small set of paths bring a significant contribution to the network's output. This set is determined by object's specificity, which sometimes can be interpretable. For example, a presence of certain objects on the picture may trigger the ResNet to transmit the information trough skip connections and not parametric transforms. This idea motivates and leads to the hypothesis that the number of required computations for inference in ResNets can be significantly reduced while saving or even increasing the efficiency of the network.

To detect this specificity, we propose a \textit{Recurrent Set (ReSet)} module, which dynamically constructs a route for each input object through layers of a trained ResNet. This route can go through a single module multiple times or skip some modules. This routing is carried out by a ReSet's part called the \textit{controller}.

Presumably, this scheme allows the network to learn a set of useful transformations and apply them in a case-specific order, that can lead to the better quality of classification and a lower number of required operations with a slightly greater number of parameters. 

While we examined only a part of these insights, we hope that this experience will be useful for the scientific community and will help researchers to further deepen their understanding of ResNet-based models properties. 

As our main contribution, we proposed the ReSet block that is indeed more flexible than SkipNet (Wang, 2017). However, SkipNet and ReSet have different motivations and outcomes—SkipNet’s controller implicitly groups objects by complexity, while our model groups objects by their semantic similarity. For example, one can use routing scores as a short set of discrete image features (or embeddings). Also, SkipNet can only skip layers and cannot add new transformations, while ReSet can learn new features by combining learned layers and increase the quality.
In our work, we described different approaches to ReSet’s construction and provided many experiments with different settings. We believe that this experience will be useful for future research on ResNets and neural networks with dynamic routing. 

In this paper, we denote our contributions as follows:
\begin{itemize}
    \item We propose a ReSet block that dynamically selects and applies transformations to the input data from a set of learnable layers (computational units).
    \item We design neural network ReSet-38 incorporating ReSet module into ResNet-38, and obtain better results on benchmarks while saving the same number of net's parameters
    \item We observed that proposed ReSet module groups objects in certain paths patterns by their semantic similarity (not only by complexity, as in described above models), which opens a wide range of possible applications of obtained path vectors (e.g. using them as image embeddings)
\end{itemize}

\section{Related Work}
Various recent studies on ResNets revealed many intriguing properties of neural networks equipped with skip connections. 

Initially proposed ResNet is organized as three groups of layers called stages. Each stage is a sequence of a few blocks of layers with a skip connection. Between the stages, a network downsamples the image to reduce its spatial dimensions.  

In addition to~\cite{veit2016residual} and~\cite{ srivastava2015highway}, authors of~\cite{greff2016highway} proposed another explanation of ResNets' robustness to deletion and reordering. Authors provided experiments showing that ResNet does not learn completely new representations within one stage, but instead its blocks refine the features extracted by the previous layer. Only after downsampling that is performed between successive stages, a new level of representation can be obtained. This process is called \textit{unrolled iterative estimation} or an \textit{iterative feature refinement}~\cite{jastrzebski2017residual}.  Later,~\cite{huang2017learning} summed up the similarities between the boosting and an iterative feature refinement.

Considered properties of residual networks were implicitly exploited by many researchers in their works, resulting in great number of modifications of ResNet, which allows to significantly reduce the number of parameters and the number of operations during inference, without a decrease of performance.

Initially,~\cite{graves2016adaptive} introduced Adaptive Computation Time (ACT) mechanism for recurrent neural networks. Since some objects are easier to classify than others, this mechanism dynamically selects the number of iterations for a recurrent neural network, while promoting the smaller number of iterations. Next,~\cite{figurnov2017spatially} developed this idea further and incorporated ACT in ResNet-based architecture, where it decides whether to evaluate or to skip certain layers. They also introduced the Spacial Adaptive Computation Time (SACT) mechanism, which applies ACT to each spatial position of the residual block. Authors reported up to 45\% reduction of the number of computations while preserving almost the same efficiency as original ResNets.

ShaResNet~\cite{boulch2017sharesnet} shares the weights of convolutional layers between residual blocks within one stage. This model is trained exactly in the same fashion as the original ResNet and reduces the number of parameters up to 40\% without a significant loss in the quality. Also,~\cite{jastrzebski2017residual} successfully reduced the number of weights three folds by sharing all residual blocks after the fifth within each stage. They investigated that sharing batch normalization statistics leads to low model efficiency, and resolved this issue by keeping a unique set of batch normalization statistics and parameters for each iteration.

SkipNet~\cite{wang2017skipnet} dynamically routes objects through a ResNet, skipping some layers. Authors developed a Hybrid Reinforcement Learning technique to reward the network for skipping blocks that have a small impact on the output. This way, they reformulated the routing problem in the context of sequential decision making, reducing the total number of computations on average by 40\%. This work is similar to ours in the idea of dynamically selecting computational units. However, SkipNet can neither change the order of blocks nor select a single block more than once. 

Finally,~\cite{leroux2018iamnn} used ideas of iterative estimation, adaptive computational time and weights sharing and proposed an Iterative and Adaptive Mobile Neural Network. The model is a compact network that consists of recursive blocks with ACT, and reduces both the size and a computational cost compared with ResNets. Authors maximally avoided redundancy using only one block, which recursively computes transformations until ACT module decides to proceed to the next stage.

\section{Learning the Routing Policy}
Following the work of~\cite{wang2017skipnet}, we formulate the routing as a policy learning problem. We propose a model that at each iteration selects a Computational Unit (CU) $F_{i_{k}}$ from a set of units $\mathcal{F} = \{F_1, \dots, F_n\}$ and applies it to the data by computing
\begin{align}
    x_{k+1} = x_k + F_{i_k}(x_k).
\end{align}
The routing policy selects a Computational Unit by calculating a distribution vector $y_k \in \mathbb{R}^n$ that assigns a probability value for the selection of each Computational Unit. Each dimension corresponds to one of $n$ available CU, where the higher value indicates the higher contribution of the corresponding unit. Then, we sample an index of a Computational Unit from the distribution $y_k$:
\begin{align}
    x^{hard}_{k+1} = x^{hard}_k + F_{i_k}(x^{hard}_k), \quad i_k \sim y_k
\end{align}

Since the selection of a computational unit is a non-differentiable function, we can consider a ``soft'' selection mechanism, that applies the policy by computing an expected output value with respect to the distribution $y_k$:
\begin{align}
\label{delta}
 x^{soft}_{k+1} = x^{soft}_k + \sum_{j=1}^n y_{k_j} F_j(x^{soft}_k).
\end{align}
Notice, that when the weight $y_{k_j}$ for a Computational Unit $F_j$ is zero, evaluation of $F_j(x_i)$ can be skipped, saving computational time. With a slight abuse of notation, we further denote $x_k^{soft}$ and $x_k^{hard}$ as $x_k$, and the type of selection mechanism that is used will be clear from the context.

We compute the weights $y_k$ using the {\it Controller} network $C$. The Controller takes a current representation of an object $x_k$ and a current state of the controller $h_k$:
\begin{align}
    [\pi_k, h_{k+1}] = C(x_k, h_k),
\end{align}
where $\pi_k$ are logits that can be transformed into a probability distribution by applying a Softmax function:
\begin{align}
    y_{k_i} = \frac{ \exp(\pi_{k_i}) }{ \sum_{j=1}^{n} \exp(\pi_{k_j})}.
\end{align}

In the general case, the soft scoring for a dense vector $y$ requires an evaluation of all CUs, requiring a lot of computational resources.

The soft selection is differentiable, but is more computationally expensive. Hard selection is a non-differentiable operation and requires some tricks to pass the gradient through sampling using REINFORCE algorithm, Gumbel trick, or Hybrid Reinforcement Learning.

\subsection{REINFORCE}
REINFORCE is a general-purpose algorithm for estimating gradient with respect to the policy~\cite{williams1992simple}, and is based on Monte-Carlo estimation of path-dependent gradients:
\begin{align}
    \nabla \mathbb{E}_{y}\mathcal{L} = \mathbb{E}_{y} \nabla \mathcal{L} \log y  \approx \frac1n \sum_{t=1}^n \nabla \mathcal{L} \log y^t, \quad y^t \sim p(y)
\end{align}

Its known issue is notoriously high variance of gradients, which makes its application difficult. Several approaches to reduce variance were made in~\cite{greensmith2004variance}. However, as experiments revealed, they appeared to be not powerful enough in the context of our problem. This result can be explained by an exponentially large number of routes that can be used for an object, requiring plenty of simulations in each pooling node for precise enough gradient estimation.

\subsection{Gumbel-Softmax Estimator}
Proposed in~\cite{jang2016categorical}, this technique performs continuous relaxation of one-hot discrete sampled vector: 
\begin{align}
y_i = \frac{ \exp{\left((\log(\pi_i) + g_i)/\tau \right) } }{  \sum_{j=1}^{n} \exp{ \left((\log(\pi_j) + g_j)/\tau\right) }},
\end{align}
where $\pi_1 \dots \pi_n$ are logits for each option, $\tau$ is a constant called a \textit{temperature}, and $g_1 \dots g_n$ are samples from a $\mathrm{Gumbel}(0, 1)$ distribution, that can be easily computed as:
\begin{align}
g_i = -\log(-\log(u_i)), \quad u_i  \sim \mathrm{Uniform}[0,1].
\end{align}
This relaxation requires a soft selection. However, when a temperature $\tau$ tends to zero, weights of all elements but one become almost one-hot, effectively selecting only a single unit.

\subsection{Straight-Through Gumbel-Softmax Estimator (ST)}
Continuous relaxations of one-hot vectors still have no sparsity (although all components except one are close to 0). To sample one-hot vector,~\cite{jang2016categorical} proposed to discretize $y$ by taking $z = \text{one-hot}(arg\,max (y))$, and use continuous approximation in the backward pass by taking $\nabla_{\theta}y \approx \nabla_{\theta}z$. Thus, gradients become biased. Nevertheless, the bias approaches zero as the temperature tends to zero. This technique was called a Straight-Through (ST) Gumbel Estimator. 

\subsection{Hybrid Reinforcement Learning}
After training the controller with sparse policy $\pi$ (i.e. only one computational unit with non-zero score on each iteration), we propose to apply Hybrid Reinforcement Learning, introduced in~\cite{wang2017skipnet}, to reduce number of CU's evaluations. For this, we add a separate CU that always returns zero. Combining with residual connection, selection of this unit will produce an identity mapping. To promote usage of this unit, we add a regularizer:
\begin{align}
  \min_{\theta, \pi} R_{\textrm{HybRL}}(\theta, \pi) = \mathbb{E}_{y}\sum_{k=1}^{t} R^{skip}_k y^{skip}_{k},
\end{align}
where $t$ is the total number of controller's invocations, $k$ is the current one, $R^{skip}_k$ is a reward for skipping the computation of a CU, i.e. using a zero unit. $R^{skip}_k$ can depend on $k$ or be a constant. $y^{skip}_k$ is an indicator of whether computation was skipped or not. Maximizing this reward promotes the network to omit evaluation of CU, reducing the number of total operations.

\section{ReSet}\label{ch:architecture}

In this section, we describe the full architecture of dynamical routing that we call a \textit{Recurrent Set} (ReSet) module. 

The model consists of a \textit{Controller} and a \textit{pool (set) of Computational Units (CUs)}. For each object, the controller assigns scores to computational units and evaluates each block with a non-zero weight. A weighted sum of the results is added to the original representation, as described in Eq.~\ref{delta}. We repeat this operation a fixed number of times, and then pass the result to next layer. 

ReSet module performs case-specific transformations of input objects, clustering their internal representations and passing them through different sets of computational units. Motivated with ResNet's robustness to deletion and shuffle of layers, the proposed recurrent structure has similarities with iterative refinement~\cite{greff2016highway}. The proposed model may include multiple evaluations of the same computational unit, giving more flexibility than iterative refinement. The model with a soft selection is evaluated as follows:
\begin{align}
\label{delta2}
 [\pi_k, h_{k+1}] = C(x_k, h_k),\\
 y_k = Softmax(\pi_k),\\
 x_{k+1} = x_k + \sum_{i=1}^n y_{k_i} F_i(x_k).
\end{align}

We also add an entropy regularizer with of two terms: promoting different routes for different objects (high entropy across cases) and and promoting deterministic routes for each object on its own (low entropy for each case):
\begin{align}
\min_{\theta} R_{ent}(\theta, y) = - \lambda_1 \sum_{i=1}^{t} \mathcal{H}\mathbb{E_{\theta}}y_i + \lambda_2 \sum_{i=1}^{t}\mathbb{E_{\theta}}\mathcal{H}y_i,
\end{align}
where $\theta$ are model parameters, and $\mathcal{H}$ is entropy.

\begin{figure*}[t]
    \centerline{\includegraphics[height=6cm]{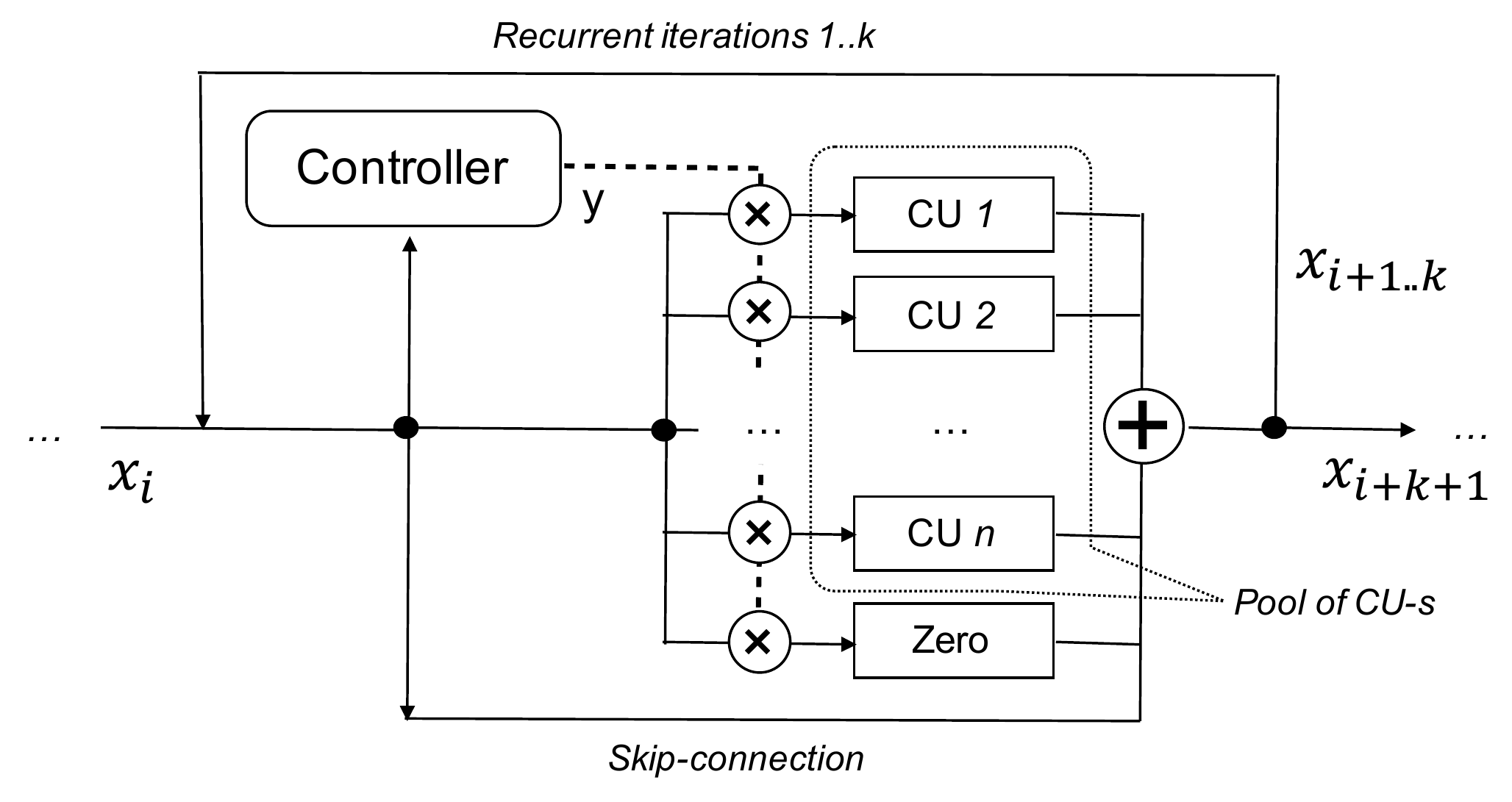}}
    \caption{ReSet module. Input $x_i$ is processed for $k$ iterations. At each iteration, either a zero function, or one of $n$ Computational Units (CU) is selected using a controller. The selected unit is evaluated with a value $x_{i+j}$, and the result is added to the current value of $x_{i+j}$, obtaining $x_{i+j+1}$. The final output of the model is $x_{i+k+1}$}
    \label{fig:reset_module}
\end{figure*}

\begin{figure}
    \subfigure[CNN controller]{\label{fig:cnn_ctr} \includegraphics[width=\fcontroller\textwidth]{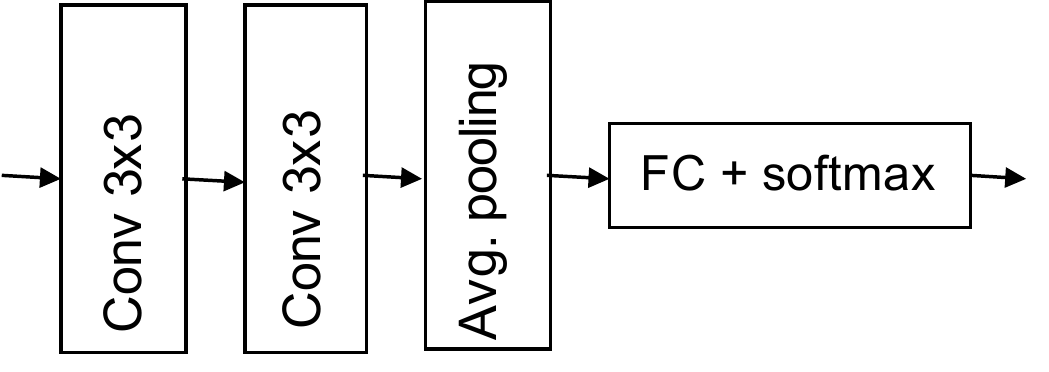}}%
    ~\subfigure[RNN controller]{\label{fig:rnn_ctr}\includegraphics[width=\fcontroller\textwidth]{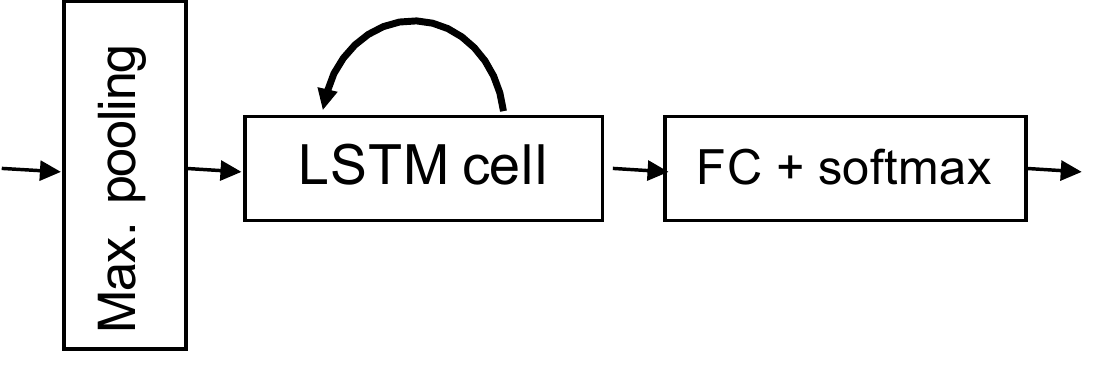}}
    \caption{Controller architectures. {\bf a)} Convolutional controller. An input to this controller is a current feature representation $x_i$.  {\bf b)} Recurrent controller. Along with $x_i$, an intermediate LSTM layer stores the hidden state between representations.}
\end{figure}

We tried two different architectures for a controller: a convolutional (CNN, Figure~\ref{fig:cnn_ctr}) and a recurrent (RNN, Figure~\ref{fig:rnn_ctr}) controller.

The \textit{CNN controller} is a shallow convolutional network without biases with a global average pooling and a fully-connected layer with a Softmax in the end. As recommended in~\cite{jastrzebski2017residual}, we store and recompute unique batch normalization statistics and parameters for each iteration. \textit{RNN controller}, uses an LSTM cell~\cite{Hochreiter1997} to perform sequential decision making within one ReSet module. The diagram of a ReSet module is shown on Figure~\ref{fig:reset_module}.

\begin{figure*}[t]
\centering
\includegraphics[width=.8\textwidth]{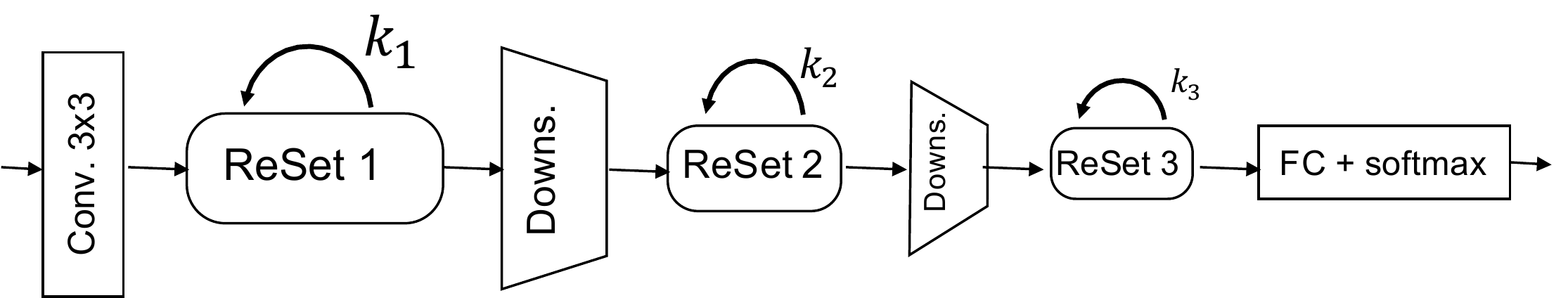}
\caption{General scheme of ResNet with dynamic recurrent routing. Three stages of ReSet with intermediate downsampling layers and a final fully-connected layers with a Softmax.}
\label{fig:general_scheme}
\end{figure*}

Further, we designed a ReSet architecture (Figure~\ref{fig:general_scheme}), that is obtained by replacing stages in ResNet-38 with ReSet modules, allowing dynamic routing through a set of Computational Units presented by ResNet blocks. The model consists of 3 sequential parts:
\begin{enumerate}
  \item A preliminary block containing a convolutional layer, followed by a batch normalization, ReLU, and a max pooling.
  \item Three ReSet modules with a similar structure. Each module contains a pool of five Computational Units $\{F_1, \dots, F_5\}$ and a zero-connection. Each CU is structured as original ResNet's block, i.e. small learnable network formed by two convolutional layers with batch normalizations and ReLU activations. We iterate each ReSet module five times with these CUs to produce the output. We also add an extra convolutional layer with a stride between the groups to downsample spatial dimensions.
  \item Final global average pooling followed by a fully-connected classifier.
\end{enumerate} 

In general, ReSet's pool of Computational Units consists of $n$ instances of a ResNet block. This decision allows to replace whole stage of original network with one ReSet module with certain number of iterations. 

The algorithm for evaluating a ReSet block is shown in Algorithm \ref{alg:reset}. The whole architecture of the proposed models is shown in Figure~\ref{fig:general_scheme}.

\begin{algorithm}[h]
    \caption{ReSet transformation}
    \label{alg:reset}
\begin{algorithmic}
    \REQUIRE ~\\ Controller $C$, Computational Units set $F$ of size $n$, representation on $i$-th level $x$, number of iterations~$k$
    \ENSURE $x_{i+k+1}$ \COMMENT{representation on $i+k+1$-th level}
    \STATE $x_i \leftarrow x$ \COMMENT{initialize result variable}
    \STATE $C.reset\_state()$ \COMMENT{reset controller's state}
    \FOR {$j \leftarrow 1, 2, \ldots, k$}
  \STATE $y_j \leftarrow C.get\_scores(x_{i+j}, j)$  \COMMENT{get scores for CUs on current iteration}
  \STATE $x_{i+j+1} \leftarrow x_{i+j} + \sum_{r=1}^{n} y_{j_r} \cdot F_r(x_{i+j})$  \COMMENT{evaluate each CU from the pool and gather results weighted by scores}
    \ENDFOR
\end{algorithmic}
\end{algorithm}

\section{Experiments}
We used CIFAR-10~\cite{krizhevsky2009learning} in provided experiments. It contains $50{,}000$ images in the training set and $10{,}000$ images in the test set, each of which has size $32\times 32$ and is related to one of $10$ classes. Also, to measure the effect of hyperparameters overfitting, we used an additional test set of images introduced in~\cite{recht2018cifar10.1, torralba2008tinyimages}.  We compare our results with the ResNet-38 architecture~\cite{he2016deep}. 

\subsection{CNN controller vs RNN controller}
To test the controller's ability to route objects, we conducted the following experiment. We took a pretrained ResNet-38 model, and put blocks from its stages into corresponding pools of computational units of three ReSet blocks. We fixed parameters of all computational units, and trained only a controller on $5{,}000$ training batches of size $128$ using Adam~\cite{kingma2014adam} optimizer with a learning rate $0.001$. We used He~\cite{he2016deep} initialization and a L2-regularizer on weights along with gradient clipping. 

In a stated scenario, a good controller should find a deterministic policy corresponding to a sequential evaluation of a ResNet. We observed a close to uniform and ineffective policies of the CNN-controller, while a RNN-controller was able to find a sequential policy.

\subsection{Learning ReSet-38 from scratch}
To validate the idea of dynamic recurrent routing, we conducted experiments with shortened ResNet, that are shown in Table~\ref{tab:short_results}. From them it can be concluded that ReSet model found effective routing and better weights compared with original ResNet.

\begin{table*}[t]
    \caption{Results on CIFAR-10 (C10) and CIFAR-10.1 (C10.1) with shortened ReSet, $n_1-n_2-n_2$ shows the number of blocks on each of 3 stages.. $\Delta$ is the performance improvement with the corresponding baseline} \label{tab:short_results}
    \centering
    \begin{tabular}{c|c|c|c|c|c|c} 
      \hline
      Arc & Model &Acc, C10&Acc, C10.1&Gap&Acc, C10 $\Delta$ & Acc, C10.1 $\Delta$ \\      
      \hline
    \multirow{2}{*}{1-1-5} & ResNet & 88.7 &\textbf{79.5} & 9.2 & - & - \\
    & ReSet & \textbf{88.9} &   \textbf{79.5} & 9.4 & 0.2  & 0 \\
    \hline
    \multirow{2}{*}{1-5-1} & ResNet & 87    &   78.1 &  8.9 & - &   - \\
    & ReSet & \textbf{88}    &  {\bf 78.7} &    9.3 &   1     & 0.6 \\
    \hline
    \multirow{2}{*}{5-1-1} & ResNet & 86.5 &    78      &  8.5 & - &    - \\
    & ReSet & \textbf{87.1}  &  {\bf 78.2} &    8.9 &   0.6 & 0.2 \\
      \hline
    \end{tabular}
\end{table*}

In the main part of experiments, we tried 3 different selection mechanisms (\textit{scorers}) for controllers: Softmax, Gumbel Softmax and Gumbel Straight-Through estimator. We tried using Adam and SGD with Momentum optimizers with stepwise learning rate decrease (Adam: multiply by 0.95 every 1000 batches, SGD: as in original paper~\cite{he2016deep}). Also, we added an entropy regularizer to promote a selection of different routes by different objects. We varied the regularizer's weight, reducing it by the end of the training. Results are presented in the first section of Table~\ref{tab:results}.

\begin{table*}[t]
\caption{
    Results on CIFAR-10 and CIFAR-10.1.
    Default optimizer is SGD, SM -- Softmax, GSM -- Gumbel Softmax, GST -- Gumbel Straight-Through estimator, ENTR -- entropy regularizer promoting selection of different policies, HybRL -- Hybrid Reinforcement Learning promoting selection of zero CUs. Contr. -- training the only controller. Column with time indicates relative difference with baseline.
    }\label{tab:results}
    \centering
    \begin{tabular}{|l|l|c|c|c|c|c|c|} 
      \hline
      \multirow{2}{*}{N} & \multirow{2}{*}{Model} &\multicolumn{2}{c|}{Accuracy}&\multirow{2}{*}{Gap}&\multirow{2}{*}{Time $\Delta$}&\multicolumn{2}{c|}{Accuracy, $\Delta$} \\
      \cline{3-4}\cline{7-8}
        & &C10 & C10.1 && &C10 & C10.1 \\
      \hline
      
      1 & \textbf{ResNet38 (baseline)}   &  \textbf{92.8}  &   83.8   &   9  &   \textbf{-}  &   \textbf{-}   &     \textbf{-}     \\
      2 & SM & 91.9   &   83.5   &  8.4   &   1.92   &   -0.9   &   -0.3  \\
      3 & SM, Adam  &   89.9   &   81.8   &   8.1   &   2  &   -2.9   &   -2 \\
      4 & \textbf{SM+ENTR}  &   \textbf{92.1}    &   \textbf{85.1}   &   \textbf{7}      &   \textbf{1.92}   &   \textbf{-0.7}   &   \textbf{1.3}    \\
      5 & SM+ENTR, Adam  &   89.3   &   79.5   &  9.8   &   2  &   -3.5   &   -4.3  \\
     6 & GSM+ENTR &   89 &   80.8   &   8.2   & 1.92   &       -3.8   & -3      \\
     7 & GST         &   60.5    &   54.1    &  6.4   & 1.08   &       -32.3 &  -29.7 \\
     \hline
     8 & SM pretr., Contr. &   91.8   & 81.9  & 9.9   & 2   &   -1   &  -1.9  \\
     9 & GSM pretr. SM, Contr. &   91.8   & 83.5  & 8.3   & 2  &     -1   & -0.3  \\
     10 & GST pretr. SM, Contr. &   66.5  & 54.1  & 12.4  & 1   &    -26.3   &  -29.7 \\
     11 & GST pretr. GSM, Contr.  &   70.1   &  57.3  & 12.8  & 1   &   -22.7    &  -26.5 \\
     12 & GST pretr. GSM     &  68.3   &    56  &  12.3 &  1    &  -24.5    &   -27.8 \\
     13 & HybRL pretr GST  & 60.7 & 51.3 &  9.4 & 0.88 &    -32.1 & -32.5 \\
     14 & HybRL pretr GST  &    48.8 &  42.2 &  6.6 & 0.79 &    -44 &   -41.6  \\
     15 & HybRL pretr GST  &    35.3 &  30.4 &  4.9 & 0.71 &    -57.5 & -53.4 \\
    \hline
     16 & 7 then finetune all & 89.8& 79.9& 9.9 &  2.08  &  -3  &    -3.9 \\
     17 & GST pretr.,  batch rep. x4 &  72.2 &  60.9 &  11.3 &  1.08  & -20.6  &    -22.9 \\
     18 & SM pretr., top-k 1     &  92.6     &  83.9       &  8.7    &  1.31   &    -0.2     &  0.1  \\
     19 & SM pretr., top-k 2    &   92.7     &  84.1    &   8.6   & 1.92  & -0.1     &  0.3  \\
     20 & SM pretr., top-k 4       &    92.1     &  83.7    &   8.4   & 1.69  & -0.7     &  -0.1 \\
      \hline
    \end{tabular}
\end{table*}

From these results we see that ReSet with a Softmax scorer and an entropy regularizer performs on-par with the original ResNet on the standard CIFAR-10 test set (used for validation here), outperforming on the CIFAR-10.1 test set. We decided to use only SGD with Momentum optimizer in further experiments because it outperformed Adam (with different hyperparameters) in almost all launches.

\subsection{Pretraining and relaxation pipelines}
Our attempts to train ReSet producing sparse scores from scratch have failed (\textit{GST} in Table~\ref{tab:results}). We assume, that this can be explained by high variance of gradients, which is due to the large number of possible controller's choices (to be precise, $5^{5
\cdot3} \approx 3\cdot10^{10}$ choices). To optimize the whole architecture, we used the following relaxation pipeline:
\begin{enumerate}
\setlength\itemsep{-0.1em}
  \item take pretrained ResNet38/ReSet38 with Softmax
  \item freeze all parameters except controller's (including batch normalization running statistics)
  \item change controller's type to Gumbel Softmax (still not sparse, but closer to Gumbel-ST)
  \item train controller on $5{,}000$ batches
  \item change controller's type to Straight-Through Gumbel Softmax
  \item train controller until policy converged
  \item add Hybrid Reinforcement Learning term to encourage reduction of operations
  \item train the network until trade-off between efficiency and performance is reached
\end{enumerate}

Results are presented in the second section of Table~\ref{tab:results}. We can see that the relaxation pipeline helps models to achieve better results compared to learning from scratch. The other important result is that while Hybrid Reinforcement Learning allows controlling computational cost, Gumbel ST estimator appeared to be too biased, leading to poor results.

\subsection{Alternative techniques}
We also tried several alternative techniques. The best-achieved results are presented in the third section of Table~\ref{tab:results}.
\subsubsection{Two-phase learning} We separated the training procedure into 2 phases: controller's and CUs' learning, and then alternated training between them. Switching from one phase to another was based on several different rules: constant number of iterations, convergence of the current phase, progressive time for one of the phases.

This strategy did not give noticeable improvements, and straightforward procedures outperformed it. We think that the main reason of this is a permanently changing loss function surface for network's parameters, which led to the divergence of the learning process.

\subsubsection{Incremental learning} Take a pretrained network, freeze all its parameters, then unfreeze and train them incrementally, starting from last layers. In other words, unfreeze last layers, train the network on some batches, unfreeze last stage, train on another set of batches, unfreeze last but one stage, and so on.

The proposed modification has shown almost the same results as two-phase learning, and our conclusions are the same too.

\subsubsection{Top-k pools choice (Softmax Straight-Through)} Compute scores with a Softmax-based controller and take the $k$ largest of them, setting others to zero. In this case, computational units are getting "unequal" updates, which leads to unstable learning. Detailed analysis and comparison with other techniques can be found in~\cite{jang2016categorical}.

This technique showed almost the same results as the baseline. Noticeably, that efficiency is not a monotonous function of $k$. That can be explained by the local properties of the gradient descent.

\subsection{Additional experiments}
We have also tried some additional experiment with computational pool's size and the number of iterations of ReSet layer. First, we used a shortened ReSet38 as a base model: we took two blocks (2 convolutions with batch normalization and ReLU) instead of the second and the third ReSet layers and modified the first ReSet layer as:
\begin{itemize}
\setlength\itemsep{-0.1em}
  \item $3$ CUs in pool and $3$ iterations (\#pool $=$ \#iters)
  \item $2$ CUs in pool and $6$ iterations (\#pool $<<$ \#iters)
  \item $2$ CUs in pool and $3$ iterations (\#pool $<$ \#iters)
  \item $6$ CUs in pool and $2$ iterations (\#pool $>>$ \#iters)
  \item $3$ CUs in pool and $2$ iterations (\#pool $>$ \#iters)
\end{itemize}

Results of the proposed modification and sequential model taken as a baseline are presented in Table~\ref{tbl:addit_results}. In this experiments, RNN-controller successfully exploits additional components and computational resources to outperform baseline model.


\begin{table*}[t]
\caption{
    Results on CIFAR-10 and CIFAR-10.1 with shortened ReSet, $n_1-n_2-n_2$ shows the number of blocks on each of 3 stages. The scorer is Softmax.
    }\label{tbl:addit_results}
    \centering
    \begin{tabular}{|l|l|c|c|c|c|c|c|} 
      \hline
      \multirow{2}{*}{N} & \multirow{2}{*}{Model} &\multicolumn{2}{c|}{Accuracy}&\multirow{2}{*}{Gap}&\multirow{2}{*}{Time $\Delta$}&\multicolumn{2}{c|}{Accuracy, $\Delta$} \\
      \cline{3-4}\cline{7-8}
        & &C10 & C10.1 && &C10 & C10.1 \\
      \hline
    1 & ResNet baseline, 2 CU \& 2 iters &  86.1 &  75      &   11.1  & -  &    -    &     - \\  
    2 & ReSet, 6 CU \& 2 iter  & 87.2 & 76.9 &  10.3 &  \textbf{1} &  1.1  &     1.9 \\
    3 & \textbf{ReSet, 3 CU \& 2 iters}   & 86.5 &  \textbf{78.4} & \textbf{8.1}   &    \textbf{1} & 0.4 &  \textbf{3.4} \\
    4 & ReSet, 2 CU \& 3 iters      &   86.4    &  77    &  9.4  &  1.08  &  0.3 &  2 \\
    5 & \textbf{SM, 2 CU \& 6 iters}          &  \textbf{87.3} &    77    & 10.3 &  1.33  & \textbf{1.2}  & 2 \\
      \hline
    \end{tabular}
\end{table*}

\subsection{Examining learned routes \label{sec:routes}}
\begin{figure}[t]
    \centering
    \includegraphics[width=\fw\columnwidth]{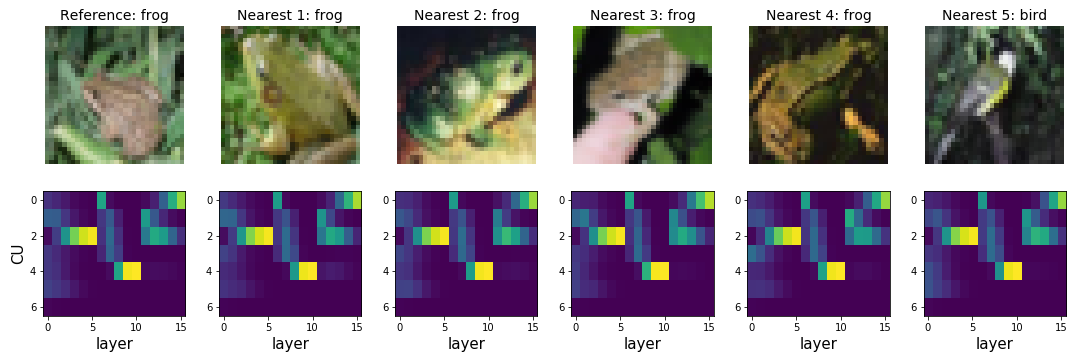}
    \includegraphics[width=\fw\columnwidth]{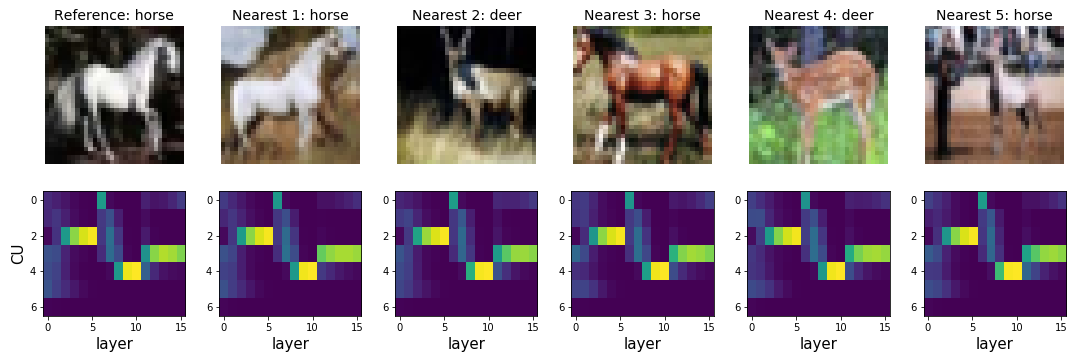}
    \includegraphics[width=\fw\columnwidth]{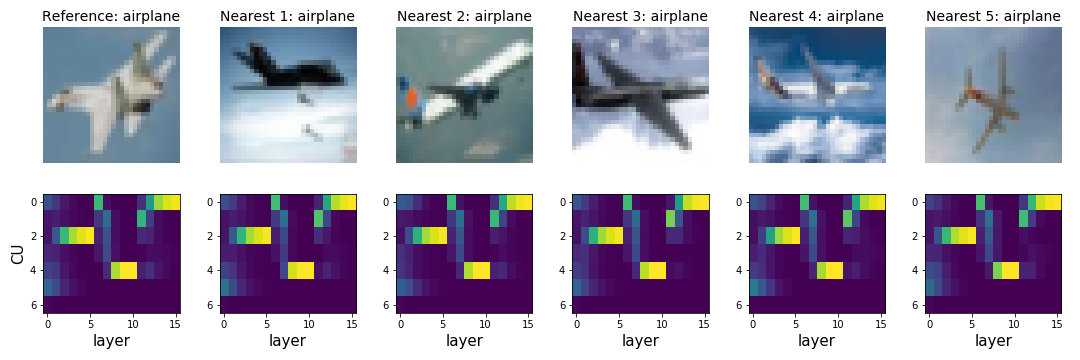}
    \includegraphics[width=\fw\columnwidth]{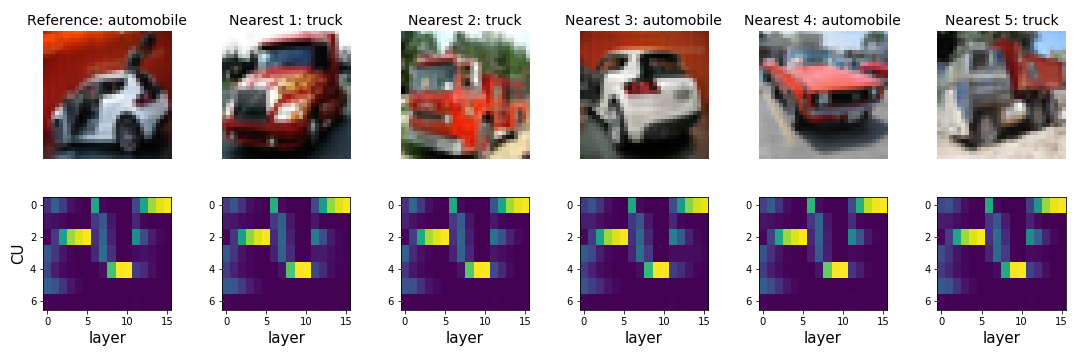}
    \includegraphics[width=\fw\columnwidth]{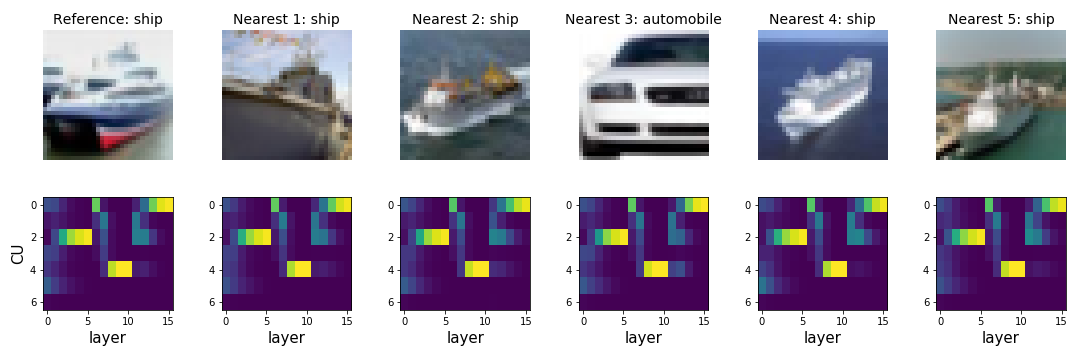}
    \caption{Routing through the network. For each experiment we show images that had the most similar routing distribution, along with the distribution itself.}
    \label{fig:route_cifar}
\end{figure}

In this section, we visualize routing scores for a few images and analyze the obtained patterns. For each considered image, we found five images with the most similar routing pattern using the Manhattan distance. The results are shown in Figure~\ref{fig:route_cifar}. On Stage 1, distribution of scores collapses---presumably, the network extracts basic patterns, and on further stages, the distribution of paths (routes) has a much higher entropy (if there is no regularizer promoting a deterministic routing).

\begin{figure}[t]
    \centering
    \hspace{-0.5cm}
    \includegraphics[width=0.5\columnwidth]{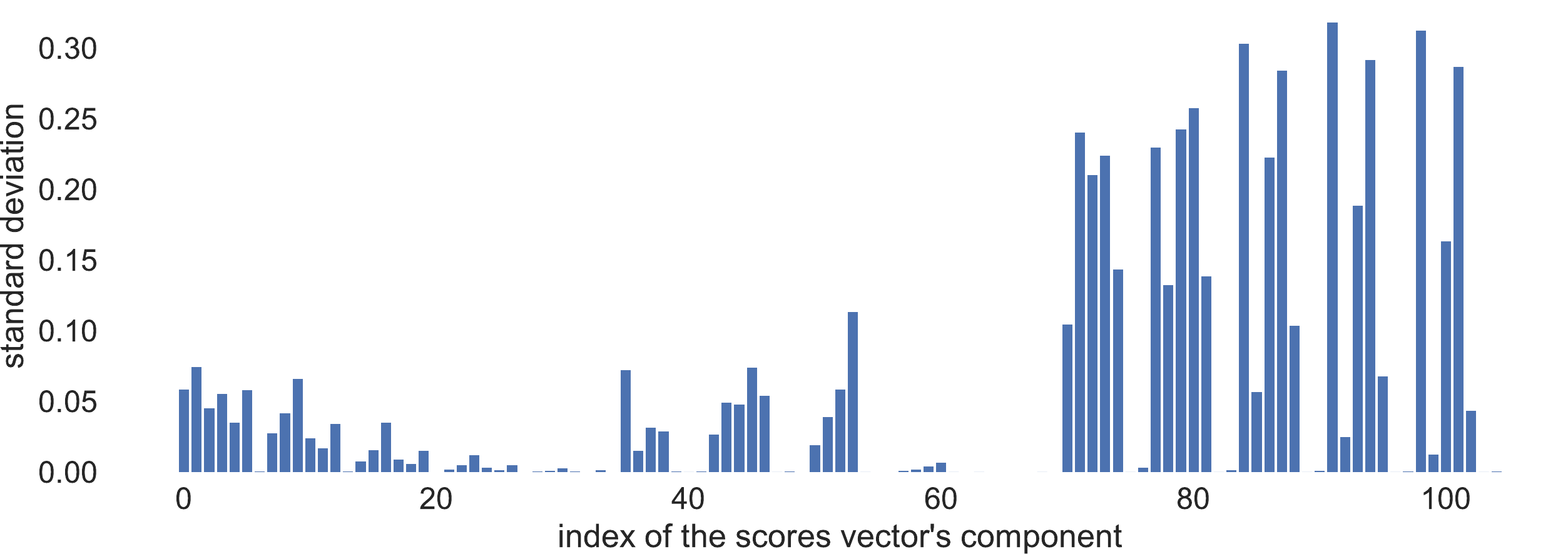}
    \includegraphics[width=0.5\columnwidth]{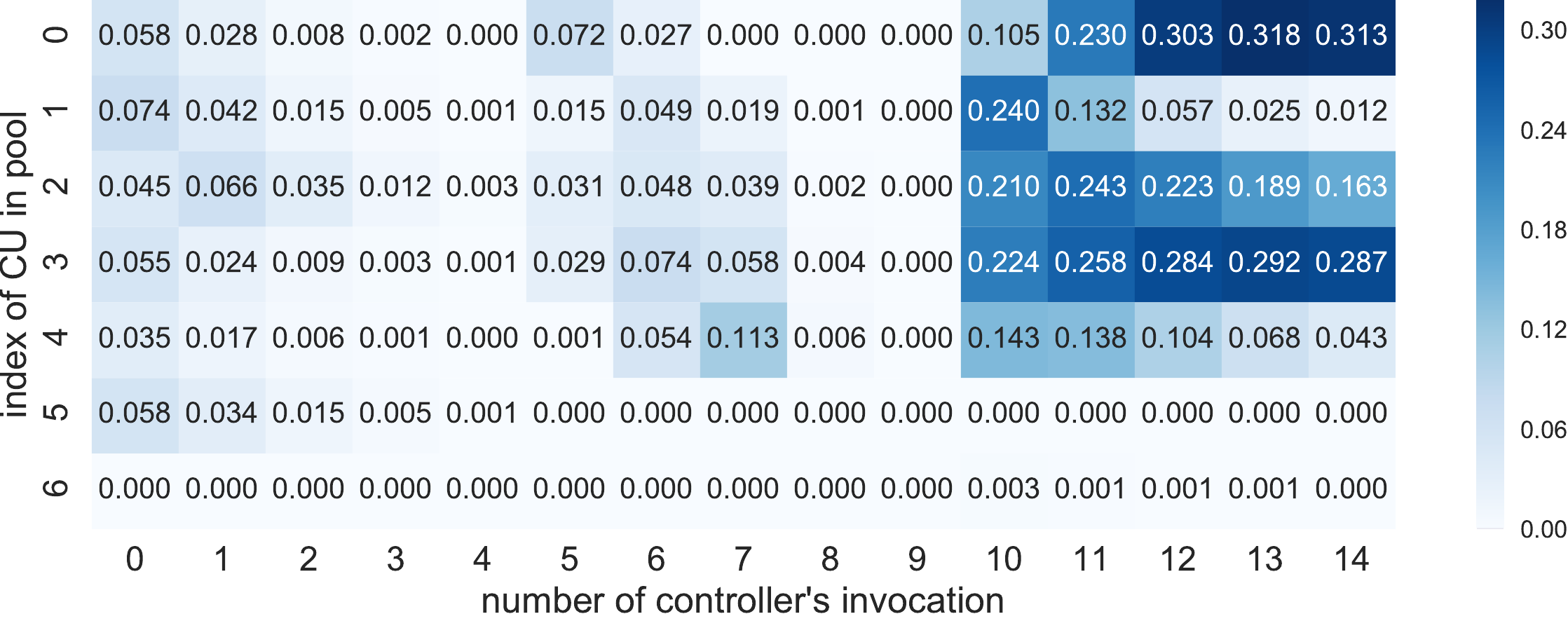}
    \caption{Standard deviations of scores vectors' components produced by model Num. 8 in Table~\ref{tab:results} for $10000$ test CIFAR-10 images. The standard deviation for certain components goes up to 0.3.}
    \label{fig:scores_std}
\end{figure}

To demonstrate this, and that routing is, indeed, dynamic, we measured standard deviations of scores vectors' components produced by model Num. 8 in Table~\ref{tab:results} for $10000$ test CIFAR-10 images. The result is shown in Figure~\ref{fig:scores_std}. 

From these figures, it implies that ReSet's controller can distinguish pictures by some intrinsic semantic properties.

Analyzing the results, it becomes apparent that the network learned to pass images of different classes through different routes where similar by L1-metric paths are assigned to semantically similar images. Also, according to obtained patterns, routing between first stages is almost identical for all objects. However, the last layers used different routes for different classes, indicating that the network uses last iterations to refine its predictions.

\subsection{Policy Evolution}
\begin{figure}[h]
    \centering
    \subfigure[Softmax-controller with entropy regularizer (Num. 3 in Table~\ref{tab:results})]{\label{fig:sm_entr_policy}\includegraphics[width=\fw\columnwidth]{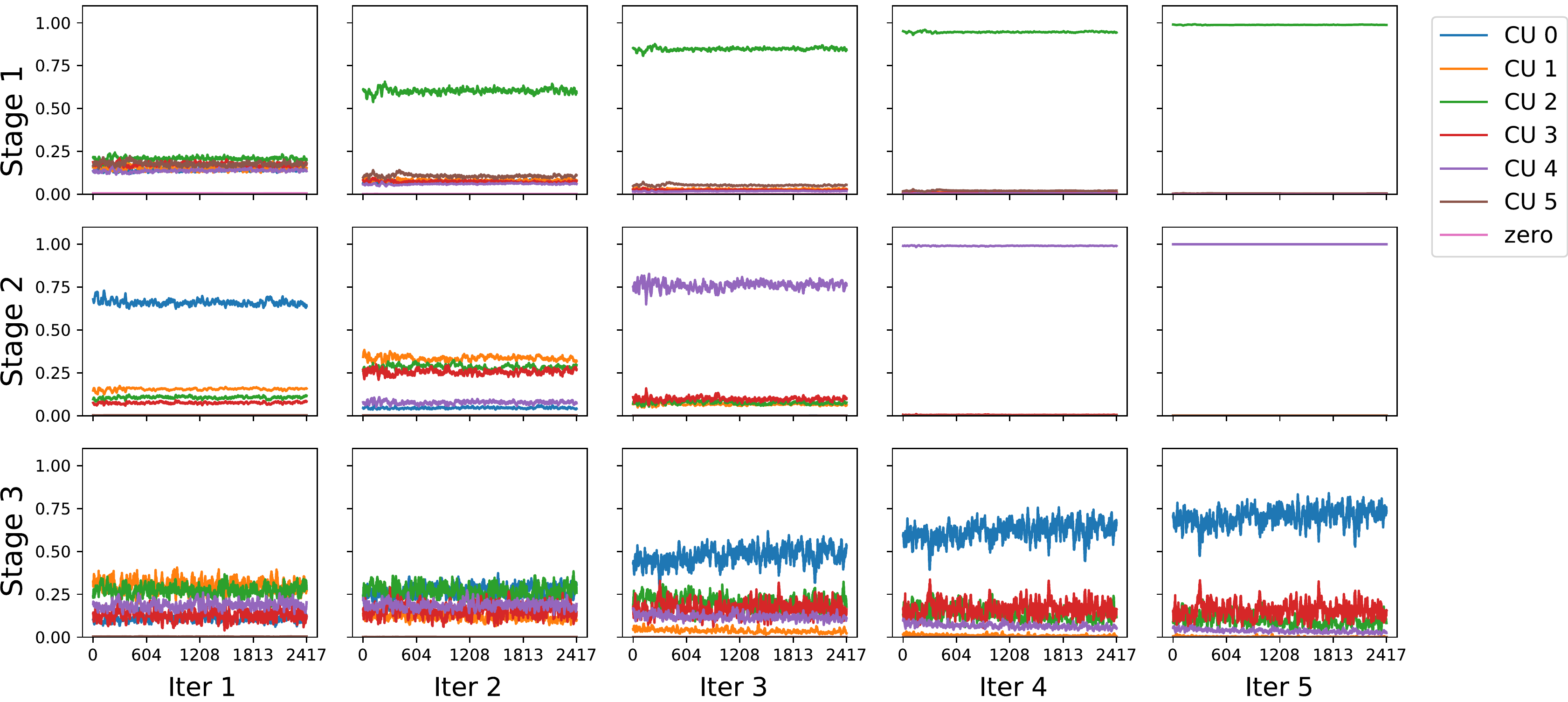}}%
    ~~\subfigure[Softmax-controller, a-priori scoring within stage]{\label{fig:sm_apriori_scores_policy}\includegraphics[width=\fw\columnwidth]{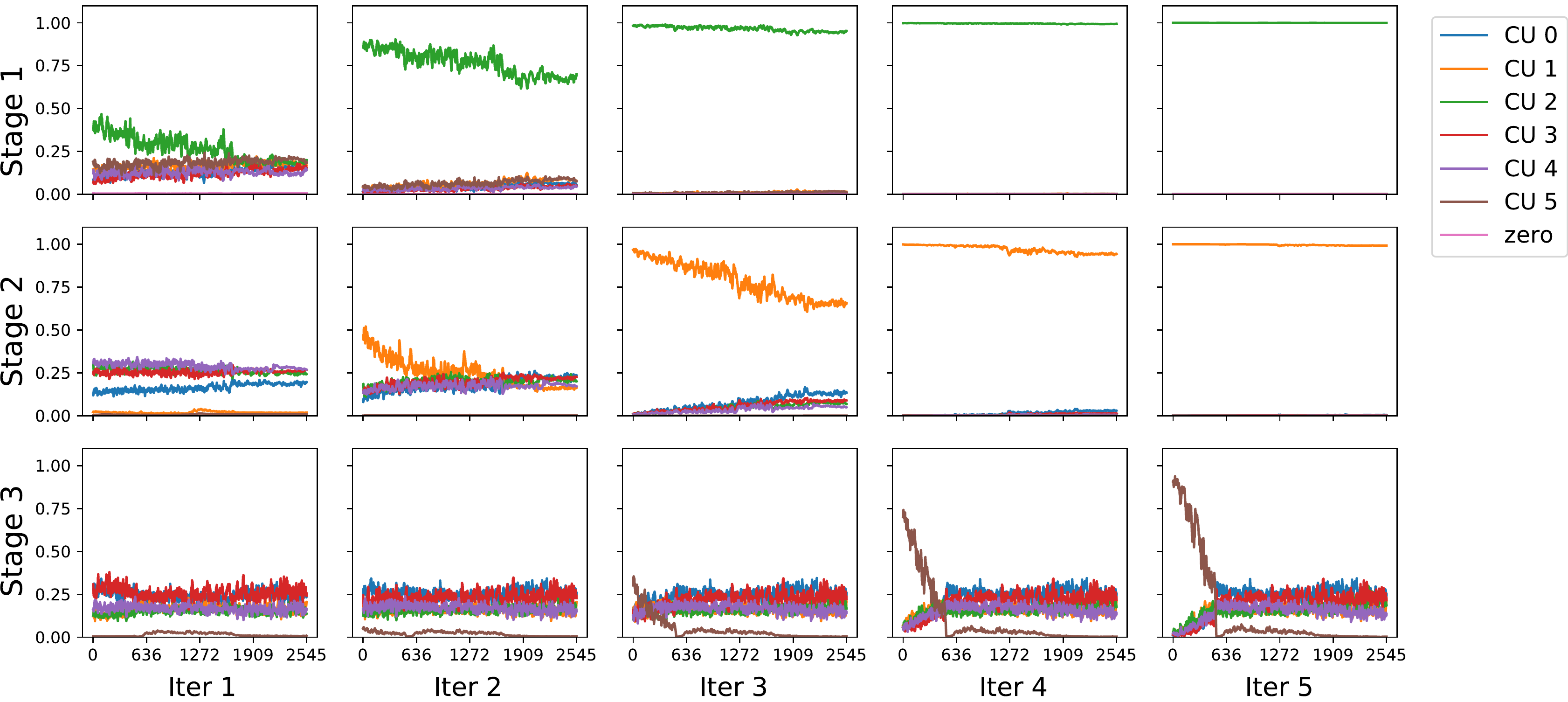}}
    
    \subfigure[Learning only controller, CU's are taken as blocks from pretrained ResNet (Num. 7 in Table~\ref{tab:results})]{\label{fig:rnn_pretrained_policy}\includegraphics[width=\fw\columnwidth]{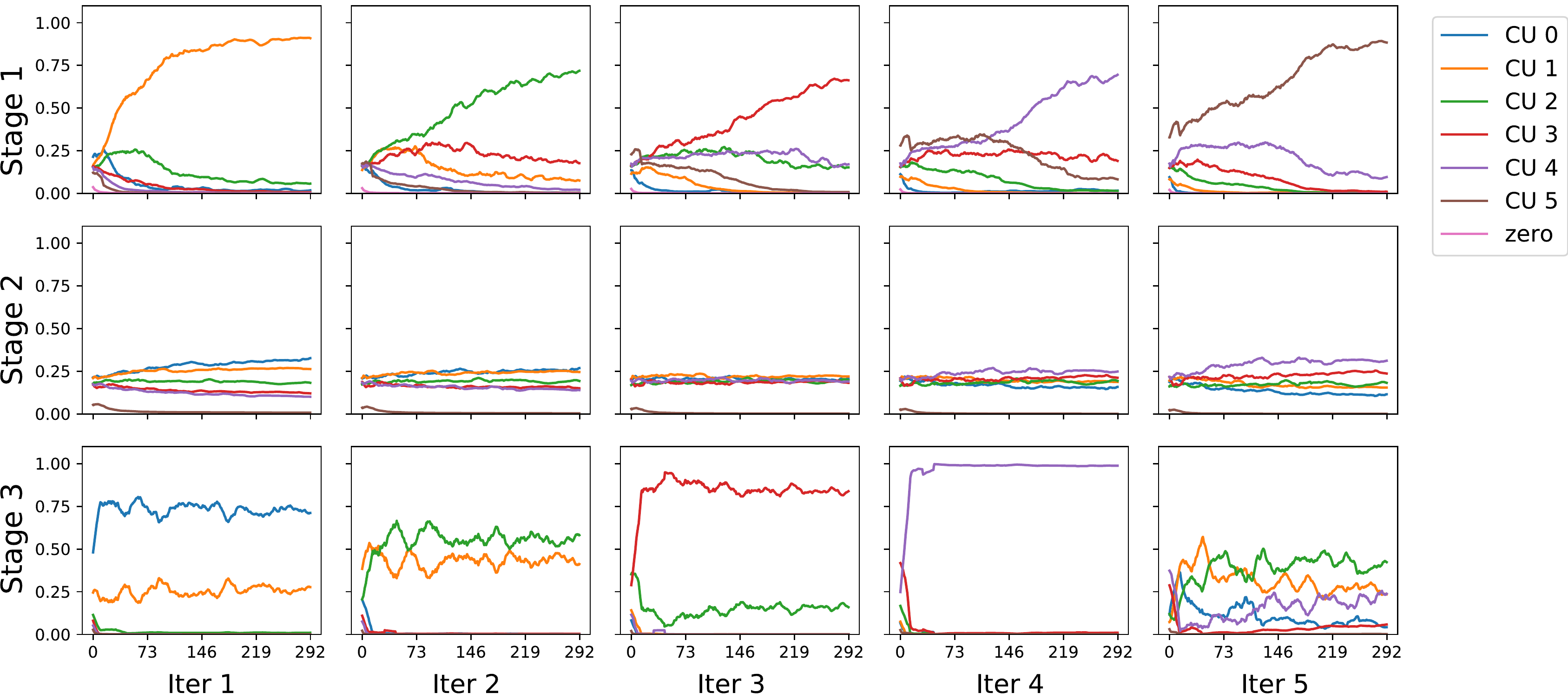}}
    ~~\subfigure[Setting as in Figure~\ref{fig:rnn_pretrained_policy} with entropy regularizer added]{\label{fig:rnn_entr_pretrained_policy}
      \includegraphics[width=\fw\columnwidth]{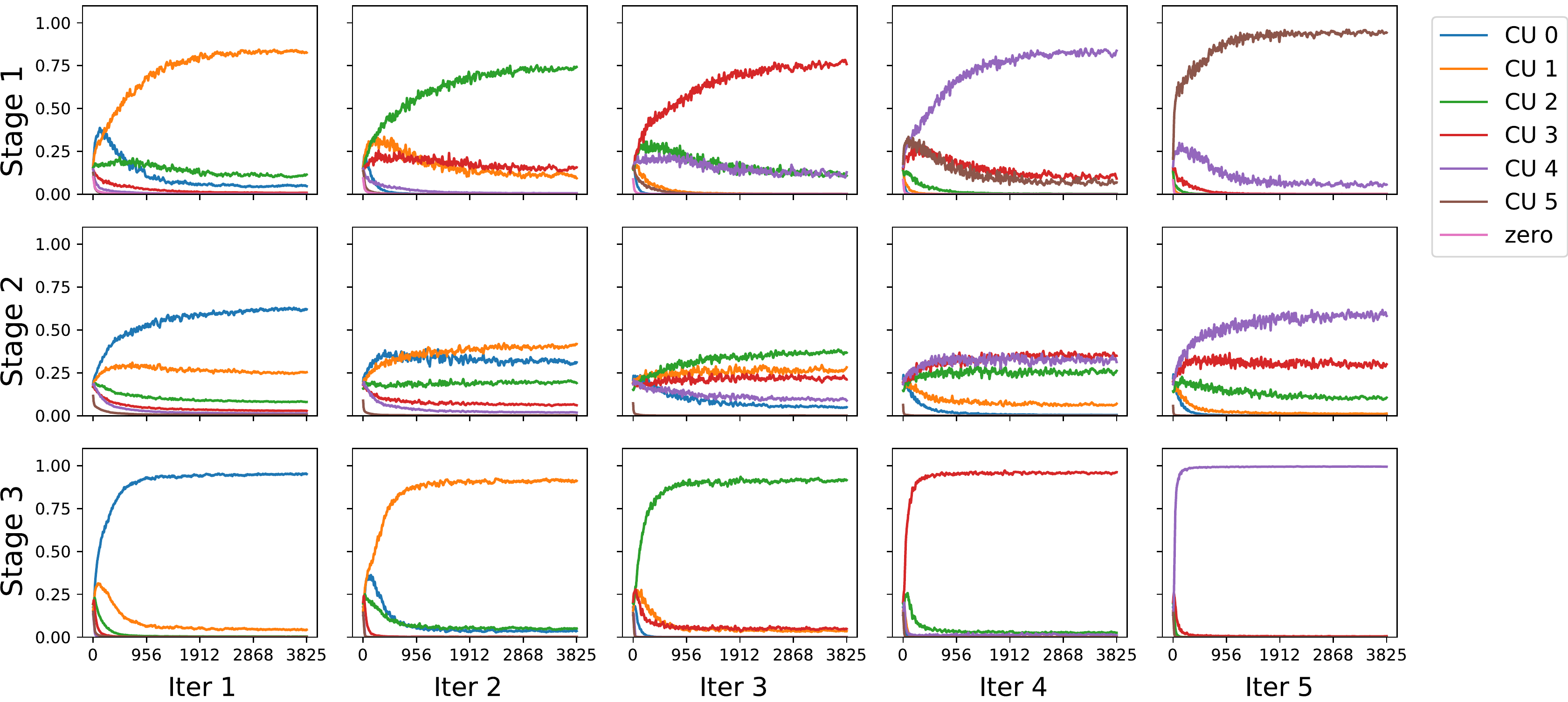}}

    \caption{Evolution of ReSet policies during training. Plots show the average score of the policy for each stage and each iteration inside the stage. An outer pair of axes describes a certain plot. In particular, an outer Y-axis shows the number of the stage number, an outer X-axis shows the number of iteration within that stage. An inner pair of axes describes the data on the plot: on inner Y-axis we placed the mean of controller's scores for certain CU, and inner X-axis shows the number of processed during training batches (in thousands, each 25-th measurement for compactness).
    }
    \label{fig:policy_evolution}
\end{figure}

In this section we visualize and analyze the evolution of policy for certain models. In  Figure~\ref{fig:policy_evolution} we show average score of different Computational Units at different stages and iterations of the ReSet model. 

From this figure, we can conclude that the learned policy can behave differently, depending on the set of the hyperparameters. For example, in Figure~\ref{fig:sm_entr_policy}, the learned policy was recurrent, since one block was selected at each iteration with almost certain probability. On stage 1, CU 2 learned to iteratively refine the result. In particular, the possibility to use a recurrent strategy was exploited in~\cite{leroux2018iamnn}. Also, some stages develop a mixed behaviour, selecting different CUs for different objects. As suggested in the previous experiment, this usually happens at the last stage.

On Figure~\ref{fig:rnn_pretrained_policy} and~\ref{fig:rnn_entr_pretrained_policy} we see that on stage 2, the policy is absolutely different from sequential (as in original ResNet-38), however, leading to the same result. This can be treated as new evidence of unrolled iterative estimation hypothesis~\cite{jastrzebski2017residual}. Noticeably, that entropy regularizer reduces the variance of scores (hence, the variance of gradients too) compared with model without it. 

\section{Conclusion}
In this work, we introduced a ReSet layer that performs dynamic routing through the set of independent Computational Units (transformations). The proposed model achieved better classification results compared with the ResNet-38 model, having a comparable number of parameters. The model learned to separate paths of images from different classes and produced separate Computational Units for the final stage of the network to refine its predictions.

The obtained results open a wide range of possible applications of the proposed mechanism of dynamic recurrent routing with ReSet. For example, ReSet could be used in Natural Language Processing, where one would expect the ReSet to process different parts of the sentence with different Computational Units. An additional direction of research is the properties of image's controller scores vectors, which could be considered as corresponding embeddings for pictures.

\acks{The work was performed according to the Russian Science Foundation Grant 17-71-20072.}

\bibliography{ms}

\end{document}